\documentclass[11pt,letterpaper]{article}

\usepackage[utf8]{inputenc}
\usepackage[T1]{fontenc}
\usepackage{amsmath,amssymb}
\usepackage{graphicx}
\usepackage[margin=1in]{geometry}
\usepackage{booktabs}
\usepackage{hyperref}
\usepackage{natbib}
\usepackage{xcolor}
\usepackage{float}
\usepackage{caption}
\usepackage{subcaption}
\usepackage{authblk}

\title{Re-analysis of the Human Transcription Factor Atlas Recovers
  TF-Specific Signatures from Pooled Single-Cell Screens with Missing Controls}

\author{%
  Arka Jain\thanks{\texttt{arkajain27@gmail.com}}%
  \and
  Umesh Sharma, MD, MBA\thanks{\texttt{sharma.umesh@mayo.edu}, Mayo Clinic}%
}

\date{}

\begin{document}
\maketitle

\begin{abstract}
Public pooled single-cell perturbation atlases are valuable resources for
studying transcription factor (TF) function, but downstream re-analysis can be
limited by incomplete deposited metadata and missing internal controls. Here we
re-analyze the human TF Atlas dataset (GSE216481), a MORF-based pooled
overexpression screen spanning 3,550 TF open reading frames and 254,519 cells,
with a reproducible pipeline for quality control, MORF barcode demultiplexing,
per-TF differential expression, and functional enrichment. From 77,018 cells in
the pooled screen, we assign 60,997 (79.2\%) to 87 TF identities. Because the
deposited barcode mapping lacks the GFP and mCherry negative controls present in
the original library, we use embryoid body (EB) cells as an external baseline
and remove shared batch/transduction artifacts by background subtraction.
This strategy recovers TF-specific signatures for 59 of 61 testable TFs,
compared with 27 detected by one-vs-rest alone, showing that robust TF-level
signal can be rescued despite missing intra-pool controls. HOPX, MAZ, PAX6,
FOS, and FEZF2 emerge as the strongest transcriptional remodelers, while
per-TF enrichment links FEZF2 to regulation of differentiation, EGR1 to Hippo
and cardiac programs, FOS to focal adhesion, and NFIC to collagen biosynthesis.
Condition-level analyses reveal convergent Wnt, neurogenic, EMT, and Hippo
signatures, and Harmony indicates minimal confounding batch effects across
pooled replicates. Our per-TF effect sizes significantly agree with Joung
et al.'s published rankings (Spearman $\rho = -0.316$, $p = 0.013$; negative
because lower rank indicates stronger effect). Together, these results show
that the deposited TF Atlas data can support validated TF-specific
transcriptional and pathway analyses when paired with principled external
controls, artifact removal, and reproducible computation.
\end{abstract}

\section{Introduction}

Transcription factors (TFs) orchestrate gene regulatory programs that determine
cell identity, proliferation, and differentiation
\citep{lambert2018human,vaquerizas2009census}. Of the approximately 1,600
human TFs, only a fraction have been functionally characterized in terms of
their downstream transcriptional effects. High-throughput perturbation screens
using single-cell RNA sequencing (scRNA-seq) have emerged as a powerful
approach to systematically profile TF function
\citep{dixit2016perturb,replogle2022mapping}.

Joung et al.\ introduced the MORF (Mammalian ORF) library system to screen
3,550 TF open reading frames via pooled lentiviral overexpression coupled with
scRNA-seq readout \citep{joung2023transcription}. Each cell in the pooled
screen receives a single TF-ORF construct bearing a unique 24-bp barcode,
enabling computational assignment of TF identity from transcriptomic data.
The resulting dataset (GEO accession GSE216481) comprises 30 samples totaling
over 250,000 cells, making it one of the largest TF perturbation atlases
available.

While Joung et al.\ focused on TF ranking and cell fate scoring, the deposited
data offers opportunities for deeper per-TF transcriptional analysis. Here we
present a systematic re-analysis using a fully automated, reproducible pipeline
that:
\begin{enumerate}
  \item Processes all 30 samples through standardized QC, normalization, and
    dimensionality reduction;
  \item Demultiplexes MORF barcodes to assign TF identity to 60,997 cells
    across 8 pooled screen samples;
  \item Performs per-TF differential expression using EB cells as a negative
    control with background subtraction (or one-vs-rest as a fallback) for
    61 TFs with sufficient cell counts;
  \item Identifies functional enrichment patterns linking TFs to biological
    pathways.
\end{enumerate}

\section{Methods}

\subsection{Data Acquisition and Preprocessing}

Raw 10x Chromium scRNA-seq data were downloaded from GEO (GSE216481).
The dataset contains 30 samples: 8 pooled MORF screen replicates
(\texttt{perturb\_S1--S8}), 4 TFv2d56 library screens, 2 differentiation
screen (DS) samples, 2 embryoid body (EB) controls, and 14 individual TF
overexpression samples (ASCL1, EOMES, GW, NFIB, PAX6, RFX4 with replicates).

Each sample was independently processed following standard single-cell
best practices \citep{luecken2019current} using Scanpy \citep{wolf2018scanpy}:
\begin{itemize}
  \item \textbf{Cell filtering}: minimum 200 genes per cell, maximum 10\%
    mitochondrial reads;
  \item \textbf{Gene filtering}: minimum 3 cells per gene;
  \item \textbf{Normalization}: library-size normalization to 10,000 counts
    per cell followed by $\log(1+x)$ transformation;
  \item \textbf{HVG selection}: top 4,000 highly variable genes using the
    Seurat v3 method \citep{stuart2019comprehensive};
  \item \textbf{Dimensionality reduction}: PCA (50 components, zero-centered
    off for sparse matrices), $k$-nearest neighbors ($k=10$), UMAP, and Leiden
    clustering.
\end{itemize}

After per-sample processing, samples were consolidated via outer join on the
gene axis (retaining the union of 19,323 genes) and re-clustered.

\subsection{MORF Barcode Demultiplexing}

The GEO deposit includes a pre-computed TF mapping file
(\texttt{GSM6681047\_180124\_TFmap.csv.gz}) containing cell barcode to TF-ORF
assignments for the 8 pooled screen samples. Each entry maps a 16-bp 10x
droplet barcode to one of: a specific TF identity (gene name and RefSeq
isoform), ``AMB'' (ambiguous, matching multiple ORFs), or ``NA'' (no barcode
detected).

We matched 10x barcodes from QC-filtered cells against the TFmap, successfully
assigning 60,997 of 77,018 cells (79.2\%) to 87 unique TF identities. An
additional 8,570 cells (11.1\%) were flagged as ambiguous and 8,451 (11.0\%)
had no detectable barcode. Ambiguous and undetected cells were excluded from
downstream per-TF analyses.

\subsection{Differential Expression Analysis}

Two complementary DE analyses were performed:

\textbf{Condition-level DE.} For samples with known identity (individual TF
overexpression lines, DS, TFv2d56), we compared each condition against the EB
control using the Wilcoxon rank-sum test. Genes with $|$log$_2$FC$|$ $>$ 1 and
Benjamini-Hochberg adjusted $p < 0.05$ were considered significant.

\textbf{Per-TF DE (pooled screen).} For the 61 TFs with $\geq$20 assigned
cells in the demuxed pooled screen, we used EB cells as a negative control.
The original MORF library includes GFP and mCherry as negative controls, and
Joung et al.\ used them in their analysis. However, these controls are absent
from the deposited TFmap file, so we cannot identify control cells within the
pool. Instead, we combined the two EB samples (18,433 cells) with the
assigned perturb cells and performed per-TF Wilcoxon rank-sum tests using EB
cells as the reference group. A relaxed threshold of $|$log$_2$FC$|$ $>$ 0.5
was used given the lower per-TF cell counts.

Because the EB and perturb samples come from different batches, raw
vs-EB comparisons conflate TF-specific effects with batch and transduction
artifacts. To isolate TF-specific signal, we applied a \textbf{background
subtraction} step: genes appearing as significant DEGs in $\geq$70\% of TF
groups (computed as $\lceil 0.70 \times 61 \rceil = 43$ TFs) were classified
as background (1,946 genes) and removed from each TF's DEG list.
As a fallback (e.g., when EB samples are unavailable), the pipeline also
supports one-vs-rest comparisons.

All DE analyses were performed using Scanpy's \texttt{rank\_genes\_groups}
\citep{wolf2018scanpy}.

\subsection{Functional Enrichment}

Over-representation analysis (ORA) was performed using Enrichr
\citep{kuleshov2016enrichr} against GO Biological Process 2023 and KEGG 2021
Human gene set libraries. Preranked Gene Set Enrichment Analysis (GSEA) was run
with 1,000 permutations using GSEApy \citep{fang2023gseapy}. Terms with
adjusted $p < 0.05$ were considered significant.

\subsection{Software and Reproducibility}

The complete pipeline is implemented in Python using Scanpy 1.10+, AnnData,
and GSEApy, with Bash orchestration scripts for end-to-end execution. All
code is available at \url{https://github.com/arkajain27/tf-atlas-signatures}. The pipeline is
idempotent: each step checks for existing outputs and skips if already
completed, enabling easy resumption across machines.

\section{Results}

\subsection{Dataset Overview and Quality Control}

After QC filtering, 254,519 cells across 30 samples were retained
after quality control. The 8 pooled screen replicates contributed 77,018
cells, while individual TF overexpression samples ranged from 3,700 to 12,000
cells each. Median genes per cell ranged from 2,500 to 4,200 across samples,
and mitochondrial content was uniformly low ($<$5\% median) after filtering.

\begin{figure}[H]
  \centering
  \begin{subfigure}[b]{0.48\textwidth}
    \includegraphics[width=\textwidth]{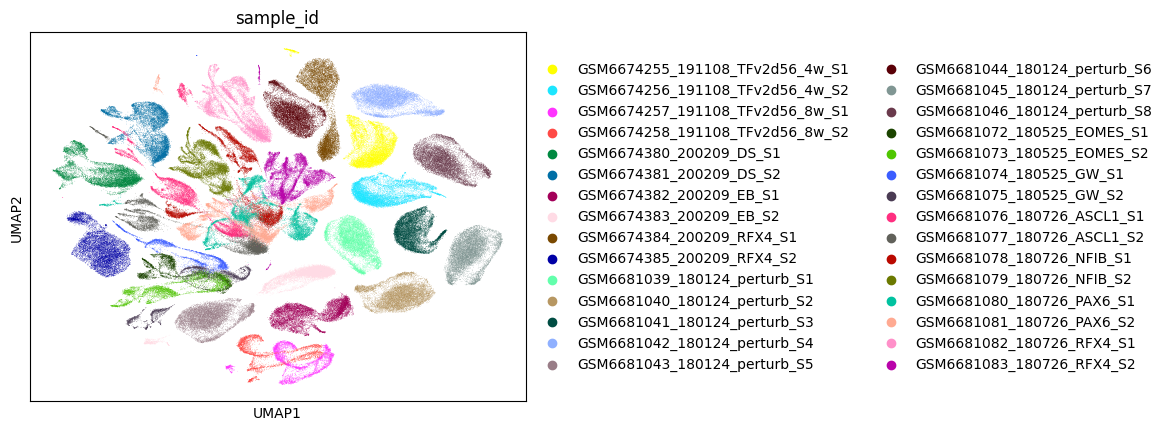}
    \caption{Colored by sample identity}
  \end{subfigure}
  \hfill
  \begin{subfigure}[b]{0.48\textwidth}
    \includegraphics[width=\textwidth]{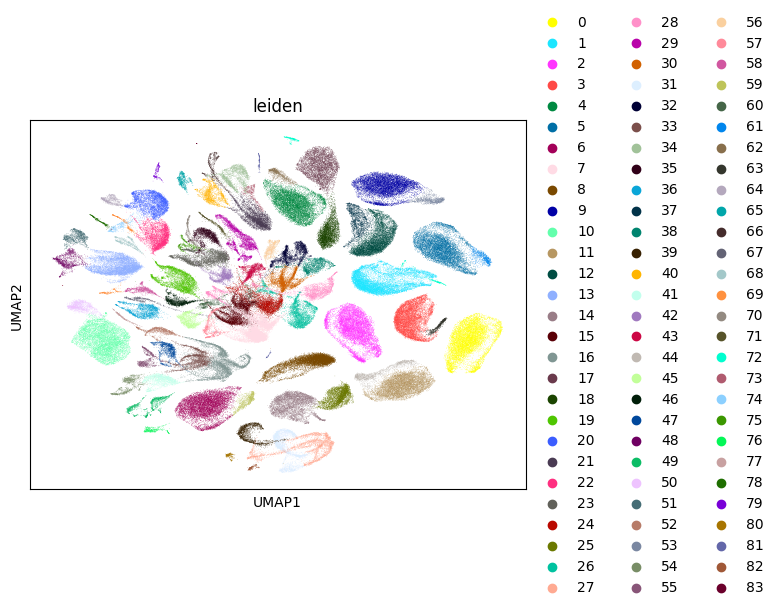}
    \caption{Colored by Leiden cluster}
  \end{subfigure}
  \caption{\textbf{UMAP of the combined TF Atlas dataset.}
    (A)~254,519 cells colored by sample identity, showing separation between
    experimental conditions and overlap among pooled screen replicates.
    (B)~Leiden clustering identifies 83 transcriptionally distinct populations.}
  \label{fig:umap_combined}
\end{figure}

UMAP visualization of the combined dataset reveals clear separation by
experimental condition (Figure~\ref{fig:umap_combined}A), with substantial
overlap among the 8 pooled screen replicates, confirming technical
reproducibility. Leiden clustering at resolution 1.0 identified 83 transcriptionally
distinct clusters (Figure~\ref{fig:umap_combined}B).

\subsection{MORF Barcode Demultiplexing}

\begin{figure}[H]
  \centering
  \begin{subfigure}[b]{0.48\textwidth}
    \includegraphics[width=\textwidth]{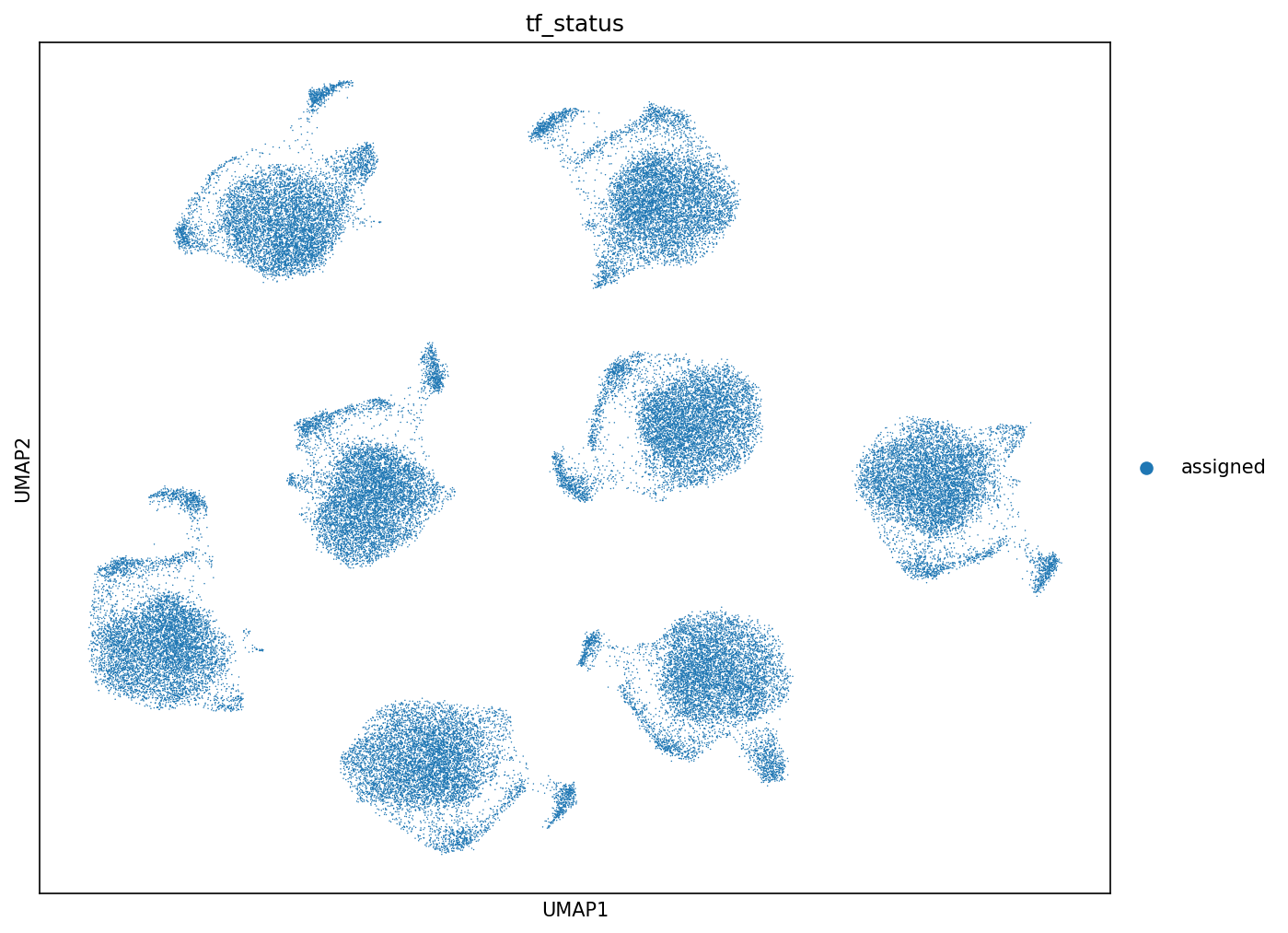}
    \caption{UMAP colored by assignment status}
  \end{subfigure}
  \hfill
  \begin{subfigure}[b]{0.48\textwidth}
    \includegraphics[width=\textwidth]{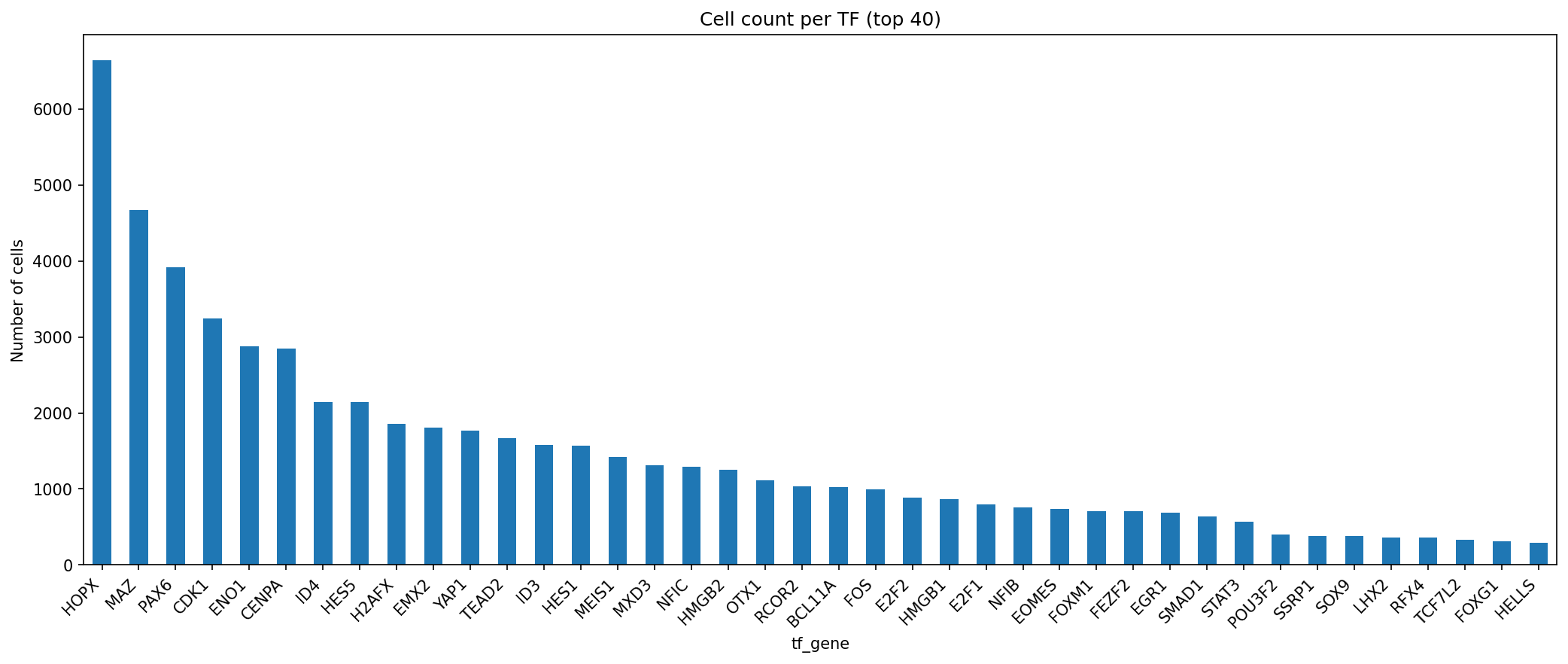}
    \caption{Cell counts per TF (top 40)}
  \end{subfigure}
  \caption{\textbf{MORF barcode demultiplexing of pooled screen samples.}
    (A)~UMAP of 77,018 cells colored by demultiplexing status: assigned (blue),
    ambiguous (orange), undetected (green).
    (B)~Distribution of assigned cells across TFs.}
  \label{fig:demux}
\end{figure}

Demultiplexing of the 8 pooled screen samples assigned 60,997 cells (79.2\%)
to specific TF identities (Figure~\ref{fig:demux}A). The assignment rate was
consistent across replicates (77.1--81.0\%). The top TFs by cell count were
HES5 (2,140 cells), ID4 (2,142), HES1 (1,573), MEIS1 (1,424), MXD3 (1,315),
and NFIC (1,294), reflecting their relative abundance in the MORF library
(Figure~\ref{fig:demux}B).

\subsection{Per-TF Differential Expression}

Using EB cells as a negative control with background subtraction, 59 of 61
TFs with $\geq$20 cells showed at least one TF-specific DEG
($|$log$_2$FC$|$ $>$ 0.5, FDR $<$ 0.05), a substantial improvement over the
one-vs-rest approach which detected only 27 TFs. The background subtraction
removed 1,946 genes that were differentially expressed in $\geq$43 of 61 TF
groups (representing batch and transduction artifacts shared across nearly all
TFs). The strongest transcriptional effects were observed for HOPX
(2,861 specific DEGs), MAZ (2,585), PAX6 (2,412), FOS (1,128), and FEZF2
(1,033) (Figure~\ref{fig:deg_per_tf}, Table~\ref{tab:top_tfs}; full results
in Supplementary Table~S1).

\begin{figure}[H]
  \centering
  \includegraphics[width=0.85\textwidth]{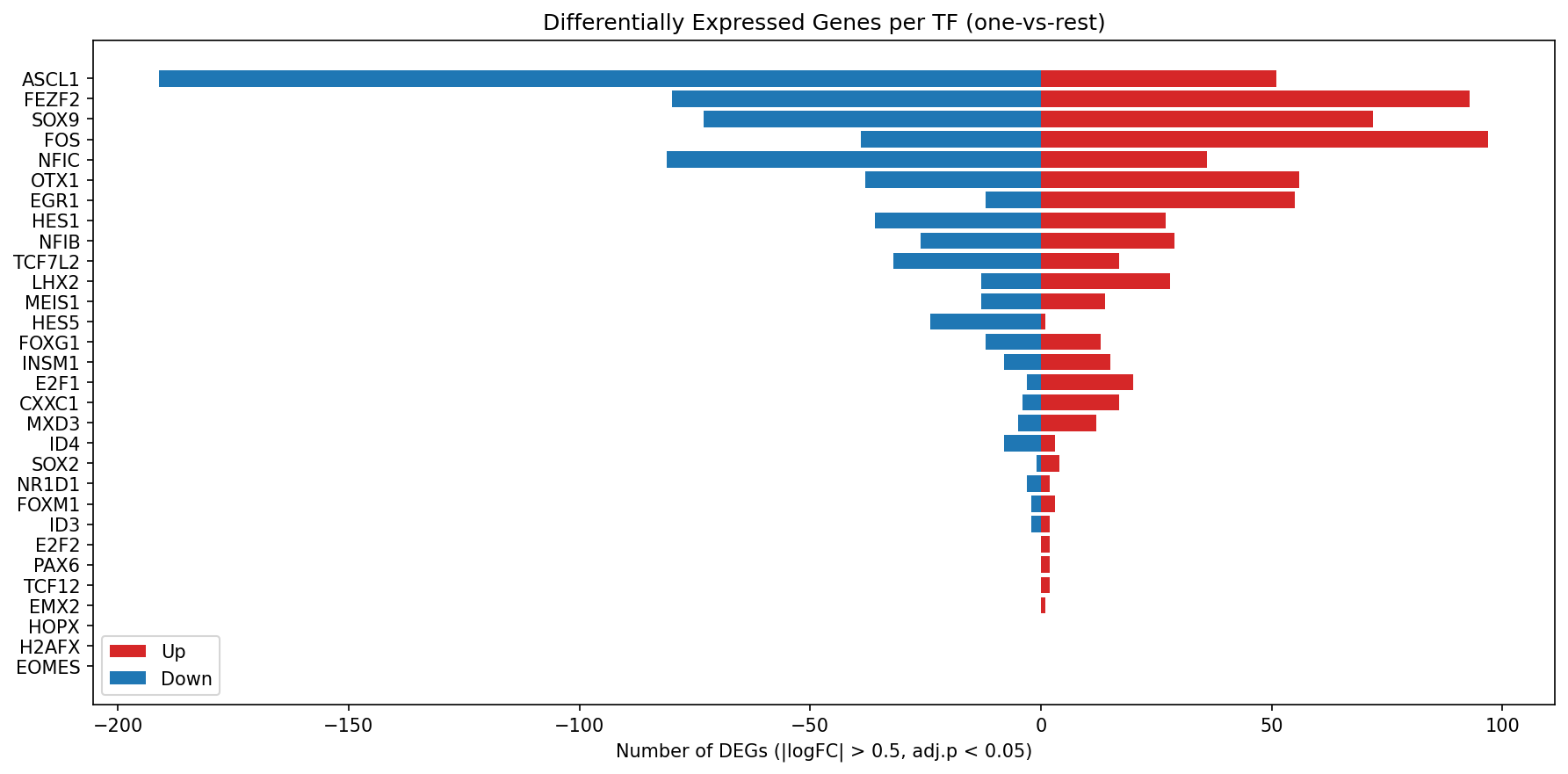}
  \caption{\textbf{TF-specific differentially expressed genes per TF} (vs EB
    control with background subtraction, $|$log$_2$FC$|$ $>$ 0.5, FDR $<$
    0.05). Red: upregulated; blue: downregulated. Top 30 TFs shown.}
  \label{fig:deg_per_tf}
\end{figure}

\begin{table}[H]
  \centering
  \caption{\textbf{Top TFs by number of TF-specific differentially
    expressed genes} from the pooled screen (vs EB control with background
    subtraction). One-vs-rest DEG counts shown for comparison.}
  \label{tab:top_tfs}
  \begin{tabular}{lrrrrrp{4cm}}
    \toprule
    TF & Cells & Specific DEGs & Up & Down & One-vs-rest DEGs & Top upregulated genes \\
    \midrule
    HOPX & 6,645 & 2,861 & 1,882 & 979 & 0 & \textit{HOPX, PAX6, SOX1, VIM} \\
    MAZ & 4,672 & 2,585 & 1,691 & 894 & 0 & \textit{MAZ, TPT1, ENO1} \\
    PAX6 & 3,913 & 2,412 & 1,569 & 843 & 0 & \textit{PAX6, SOX1, FGF8} \\
    FOS & 990 & 1,128 & 702 & 426 & 136 & \textit{ANXA1, CDH13, UTF1, EGR1} \\
    FEZF2 & 703 & 1,033 & 680 & 353 & 173 & \textit{FOXA2, TFF3, FOXB1, SOX21} \\
    NFIC & 1,294 & 914 & 553 & 361 & 117 & \textit{NUPR1, S100A14, ACTG2} \\
    ASCL1 & 155 & 43 & 29 & 14 & 242 & \textit{TFF3, FOXA2, CALCA} \\
    \bottomrule
  \end{tabular}
\end{table}

The background subtraction dramatically changes the ranking of TFs. HOPX,
MAZ, and PAX6---which showed zero DEGs in the one-vs-rest analysis---emerge
as the top three transcriptional remodelers, consistent with their high cell
counts and known biological functions. Conversely, ASCL1 drops from 242
one-vs-rest DEGs to 43 TF-specific DEGs after removing the 1,946 background
genes, indicating that much of its apparent signature was shared transduction
artifact rather than TF-specific signal. The background-subtracted ASCL1
results recover known neuronal targets (TFF3, FOXA2, CALCA) without the
ribosomal and housekeeping gene contamination that dominated the one-vs-rest
output.

FEZF2, a key regulator of corticofugal projection neuron identity
\citep{chen2005fezl}, showed 1,033 specific DEGs (680 up, 353 down) with 703
cells.

\subsection{Condition-Level Differential Expression}

Comparing individual TF overexpression samples against EB controls revealed
substantially larger transcriptional effects, with 1,910--5,641 DEGs per
condition ($|$log$_2$FC$|$ $>$ 1, FDR $<$ 0.05; Supplementary Tables~S3, S4). The pooled \texttt{perturb}
samples collectively showed 5,641 DEGs (3,882 up, 1,759 down), reflecting the
mixed population of TF-expressing cells.

Among individual TFs, EOMES showed the strongest effect (3,685 DEGs),
consistent with its central role in mesendoderm specification. ASCL1 yielded
3,338 DEGs at the condition level compared to 242 in the one-vs-rest pooled
analysis, highlighting the power of dedicated overexpression experiments versus
pooled screens for detecting subtle expression changes.

\subsection{Pathway Enrichment}

\begin{figure}[H]
  \centering
  \begin{subfigure}[b]{0.48\textwidth}
    \includegraphics[width=\textwidth]{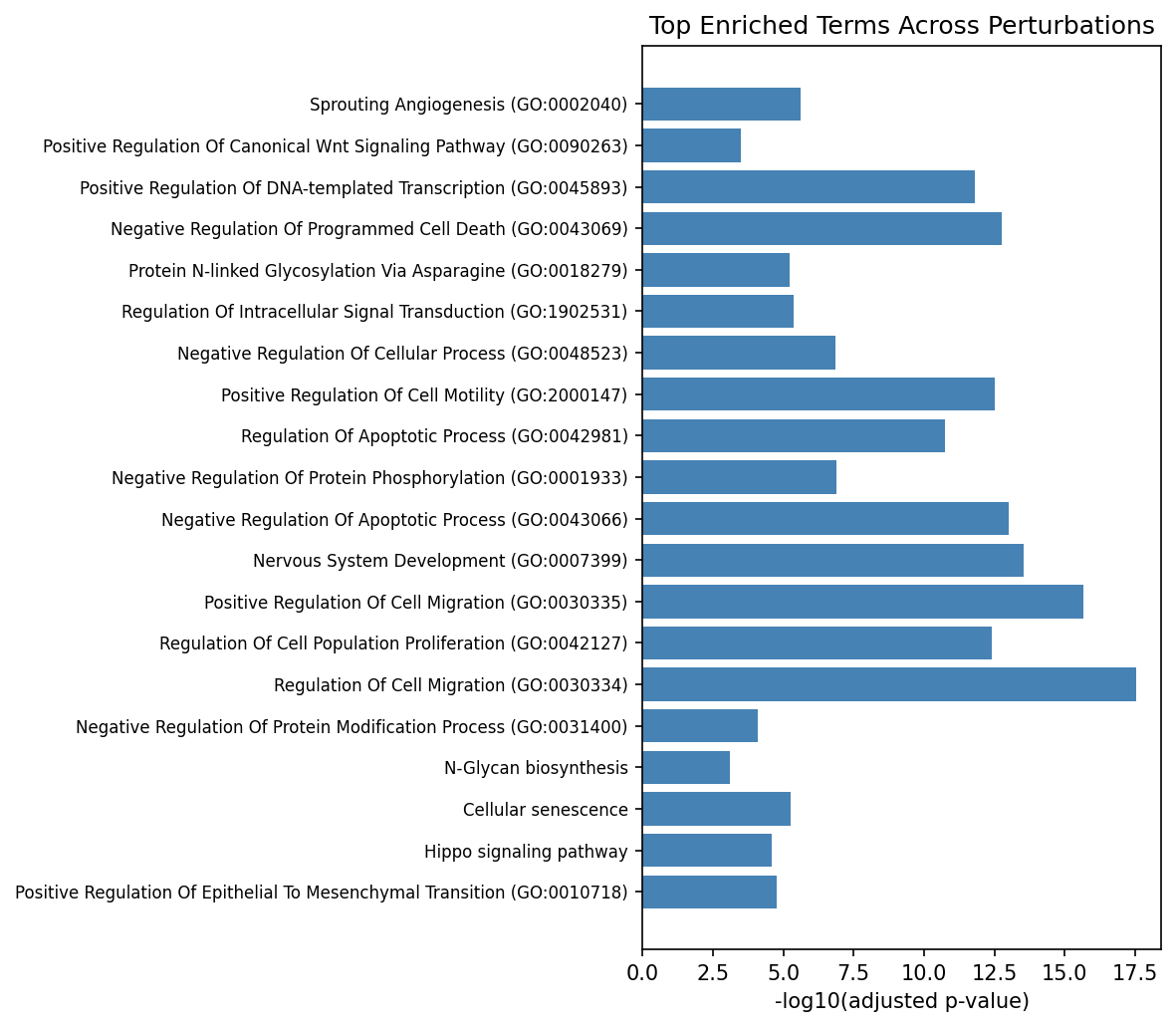}
    \caption{Top enriched terms}
  \end{subfigure}
  \hfill
  \begin{subfigure}[b]{0.48\textwidth}
    \includegraphics[width=\textwidth]{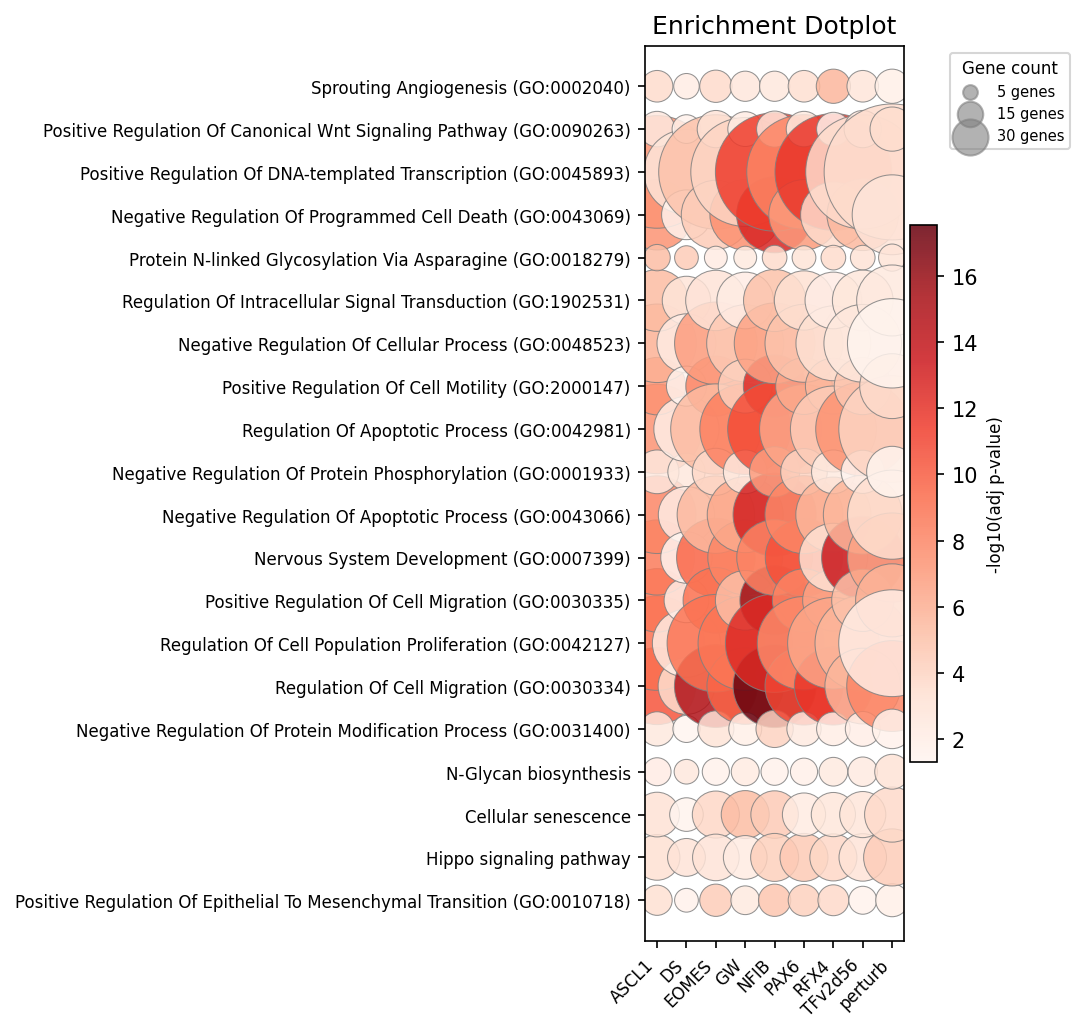}
    \caption{Enrichment dotplot}
  \end{subfigure}
  \caption{\textbf{Functional enrichment of condition-level DEGs.}
    (A)~Most frequently enriched GO/KEGG terms across perturbations, ranked by
    minimum adjusted $p$-value.
    (B)~Dotplot showing term enrichment across perturbations (dot size =
    gene count, color = $-$log$_{10}$ adjusted $p$-value).}
  \label{fig:enrichment}
\end{figure}

Enrichment analysis of condition-level DEGs revealed convergent pathway
activation across multiple TFs (Figure~\ref{fig:enrichment}). The most
frequently enriched pathways included:

\begin{itemize}
  \item \textbf{Wnt signaling} (enriched in ASCL1, EOMES, RFX4, TFv2d56):
    consistent with Wnt's role in neural patterning and stem cell maintenance;
  \item \textbf{Neurogenesis and axon guidance} (ASCL1, PAX6, NFIB, FEZF2):
    reflecting the neural differentiation context of the screen;
  \item \textbf{Epithelial-mesenchymal transition} (EOMES, DS):
    indicating cell migration programs;
  \item \textbf{Hippo/YAP signaling} (FOS, SOX9, TFv2d56):
    suggesting mechanotransduction pathway crosstalk.
\end{itemize}

\subsection{Per-TF Pathway Enrichment from Pooled Screen}

\begin{figure}[H]
  \centering
  \begin{subfigure}[b]{0.48\textwidth}
    \includegraphics[width=\textwidth]{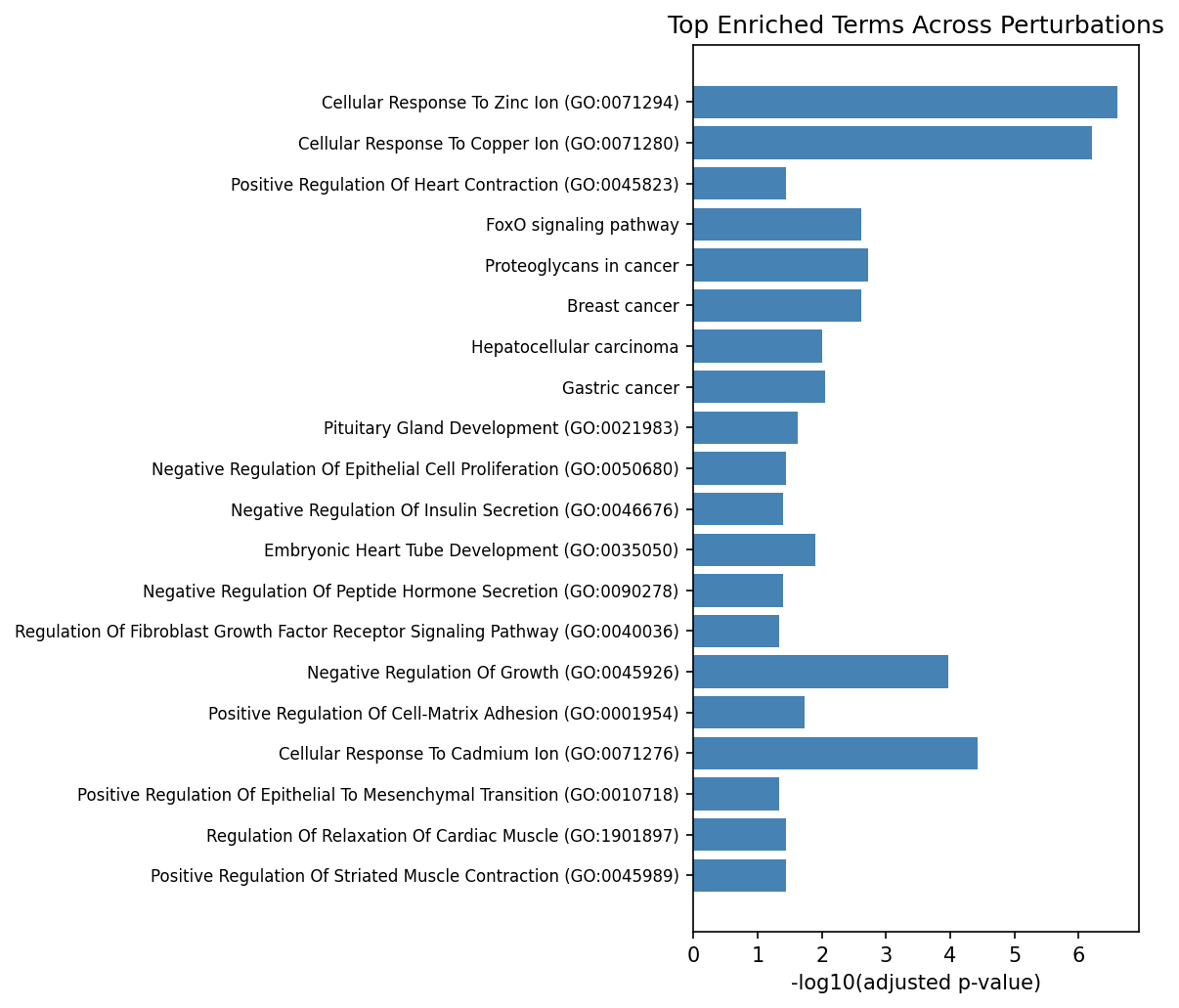}
    \caption{Top enriched terms across TFs}
  \end{subfigure}
  \hfill
  \begin{subfigure}[b]{0.48\textwidth}
    \includegraphics[width=\textwidth]{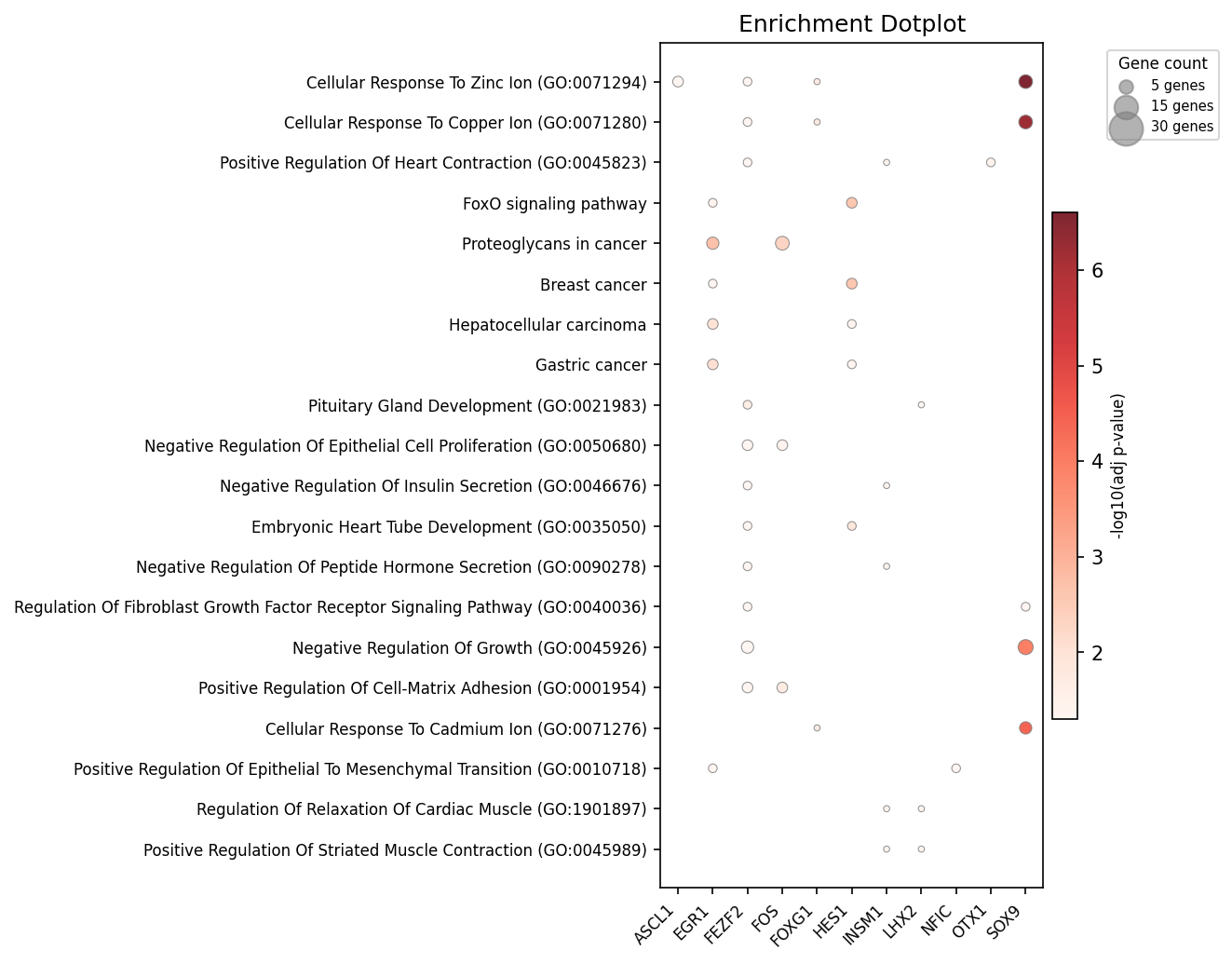}
    \caption{Per-TF enrichment dotplot}
  \end{subfigure}
  \caption{\textbf{Functional enrichment of per-TF DEGs from pooled screen.}
    (A)~Most frequently enriched GO/KEGG terms across 13 TFs with significant
    ORA results, ranked by minimum adjusted $p$-value.
    (B)~Dotplot showing per-TF enrichment landscape (dot size = gene count,
    color = $-$log$_{10}$ adjusted $p$-value). Only TFs with $\geq$5 significant
    DEGs ($|$log$_2$FC$|$ $>$ 0.5, FDR $<$ 0.05) were tested.}
  \label{fig:pertf_enrichment}
\end{figure}

To characterize the functional programs driven by individual TFs in the pooled
screen, we performed over-representation analysis (ORA) on the per-TF DEG lists.
Preranked GSEA was not applied at this level because the EB-specific DEG tables
are pre-filtered (background-subtracted), making ranked enrichment on truncated
input inappropriate. Of the
61 TFs tested, 13 yielded significant ORA terms (adjusted $p < 0.05$),
totaling 207 enriched GO/KEGG terms (Figure~\ref{fig:pertf_enrichment};
Supplementary Table~S2).

The per-TF enrichment revealed TF-specific pathway signatures consistent with
known biology:
\begin{itemize}
  \item \textbf{FEZF2} (39 terms): negative regulation of cell
    differentiation (GO:0045596) and neuron differentiation (GO:0045665),
    consistent with its role in maintaining progenitor identity during
    corticofugal neuron specification;
  \item \textbf{EGR1} (30 terms): ventricular cardiac muscle development
    (GO:0003229) and Hippo signaling pathway, reflecting its role as an
    immediate-early TF in mechanosensitive programs;
  \item \textbf{FOS} (22 terms): focal adhesion (KEGG) and proteoglycans in
    cancer, consistent with AP-1 complex regulation of cell adhesion and
    extracellular matrix remodeling;
  \item \textbf{SOX9} (17 terms): cellular response to zinc and copper ions
    (GO:0071294, GO:0071280), linking SOX9 to metal ion homeostasis in
    addition to its canonical chondrogenic role;
  \item \textbf{NFIC} (13 terms): positive regulation of collagen biosynthetic
    process (GO:0032967), consistent with NFI family involvement in connective
    tissue gene regulation;
  \item \textbf{OTX1} (3 terms): brain development (GO:0007420) and CNS
    development (GO:0007417), directly reflecting its known function in
    anterior brain patterning.
\end{itemize}

\begin{figure}[H]
  \centering
  \includegraphics[width=0.95\textwidth]{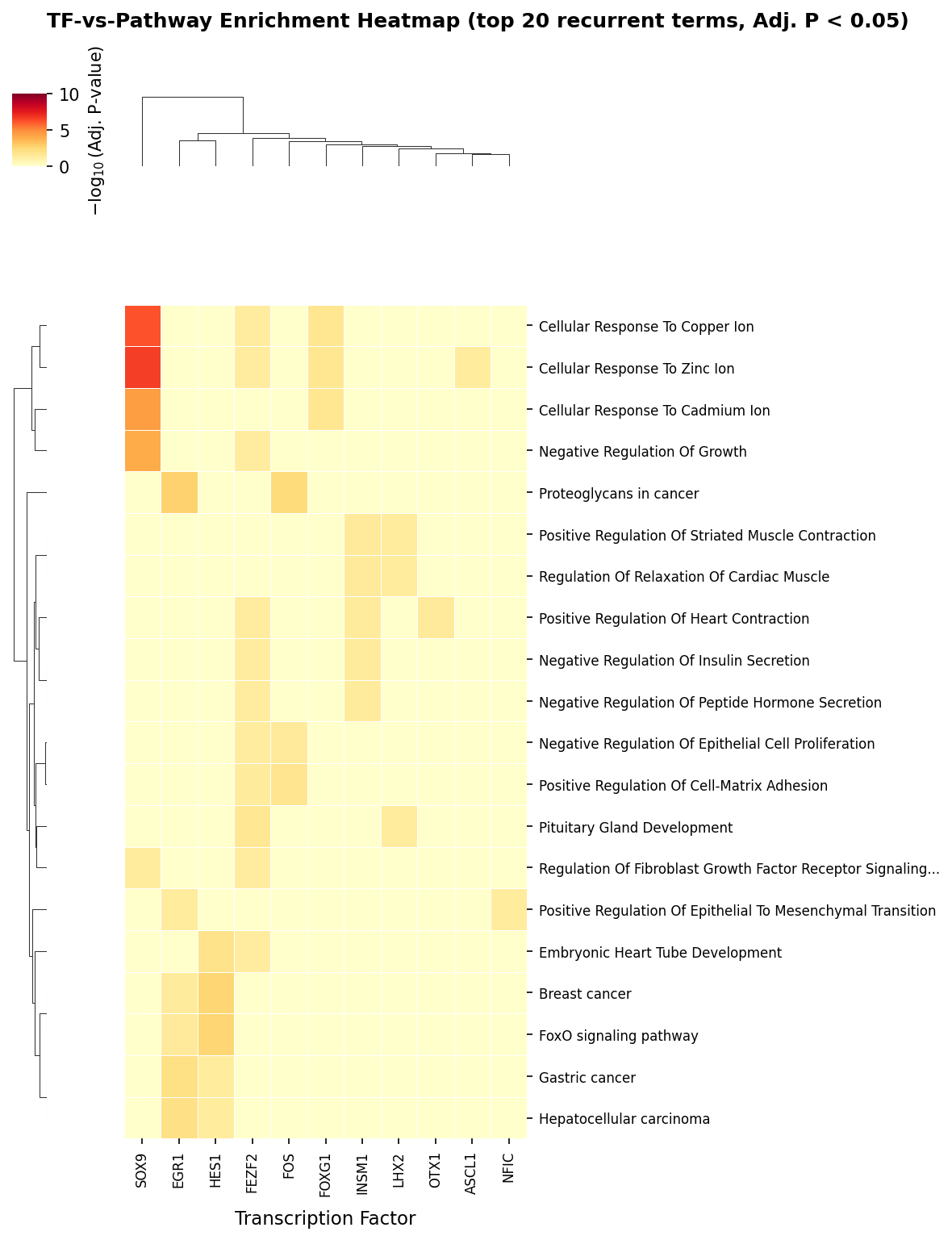}
  \caption{\textbf{TF-vs-pathway enrichment heatmap.}
    Clustered heatmap of the top 20 most recurrent enriched terms across 11 TFs
    with significant ORA results. Color intensity represents $-$log$_{10}$
    (adjusted $p$-value), clipped at 10. Hierarchical clustering reveals
    modules of co-enriched pathways and TFs with similar functional programs.}
  \label{fig:heatmap}
\end{figure}

The most recurrently enriched terms across TFs included cellular response to
zinc ion (4 TFs), copper ion response (3 TFs), and positive regulation of
heart contraction (3 TFs), suggesting shared metal ion signaling and cardiac
differentiation programs among the screened TFs. A clustered heatmap of TF-
vs-pathway enrichment (Figure~\ref{fig:heatmap}) reveals distinct modules:
a metal ion response cluster (SOX9, NFIB, FEZF2), a cardiac/muscle
contraction cluster (EGR1, HES1, OTX1), and an adhesion/EMT module (FOS,
NFIC).

\subsection{Batch Effects Assessment}

\begin{figure}[H]
  \centering
  \includegraphics[width=0.95\textwidth]{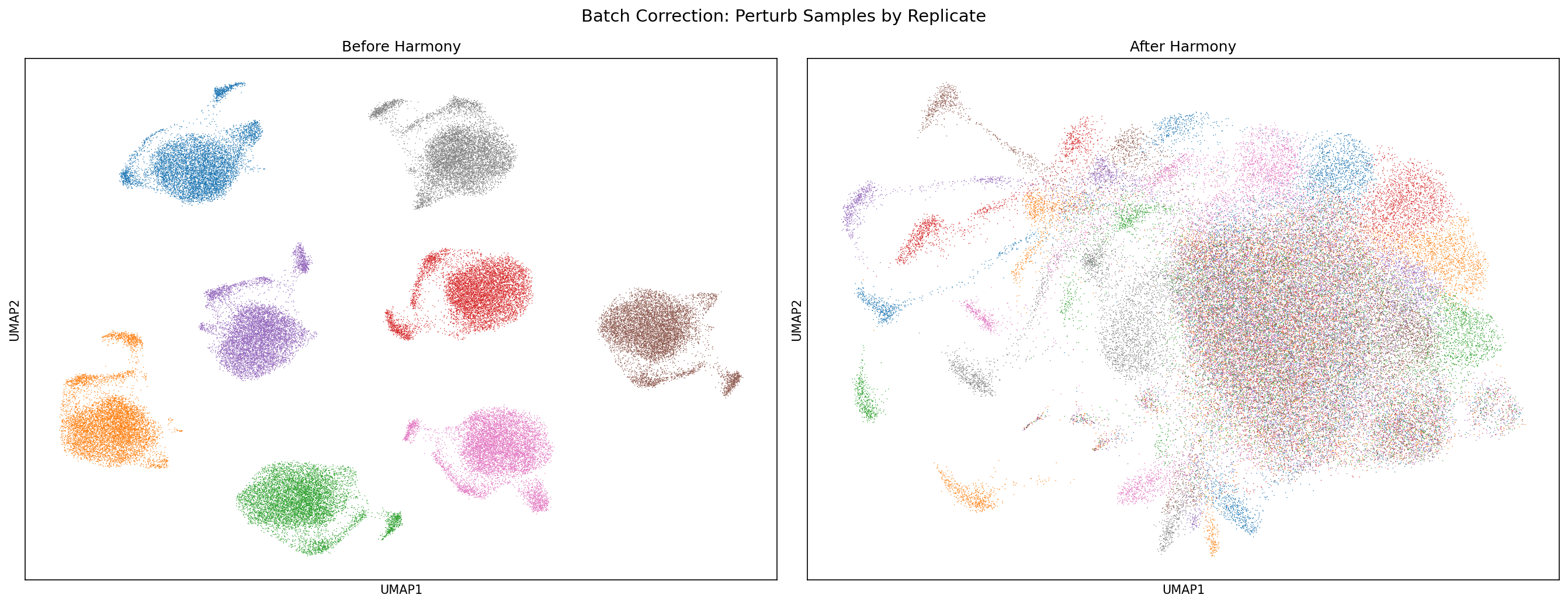}
  \caption{\textbf{Batch effects assessment via Harmony integration.}
    UMAP visualizations of the 8 pooled screen replicates before and after
    Harmony batch correction. Minimal changes confirm that inter-replicate
    batch effects are negligible and do not confound downstream analyses.
    Harmony converged in 3 iterations.}
  \label{fig:batch_correction}
\end{figure}

To assess potential confounding by technical batch effects across the 8 pooled
screen replicates, we applied Harmony integration \citep{korsunsky2019fast}
on the PCA embedding. Harmony converged in just 3 iterations (out of a maximum
20), and the before/after UMAP visualizations showed only minor differences
(Figure~\ref{fig:batch_correction}).
The per-replicate demultiplexing assignment rates were highly consistent
(77--81\%). Since Harmony corrects the PCA embedding rather than the raw count
matrix, it does not affect DE results. We therefore do not apply batch
correction to the data used for DE or enrichment analysis; the Harmony result
is retained solely as a verification that batch effects are negligible.

\subsection{Validation Against Published TF Rankings}

\begin{figure}[H]
  \centering
  \includegraphics[width=0.95\textwidth]{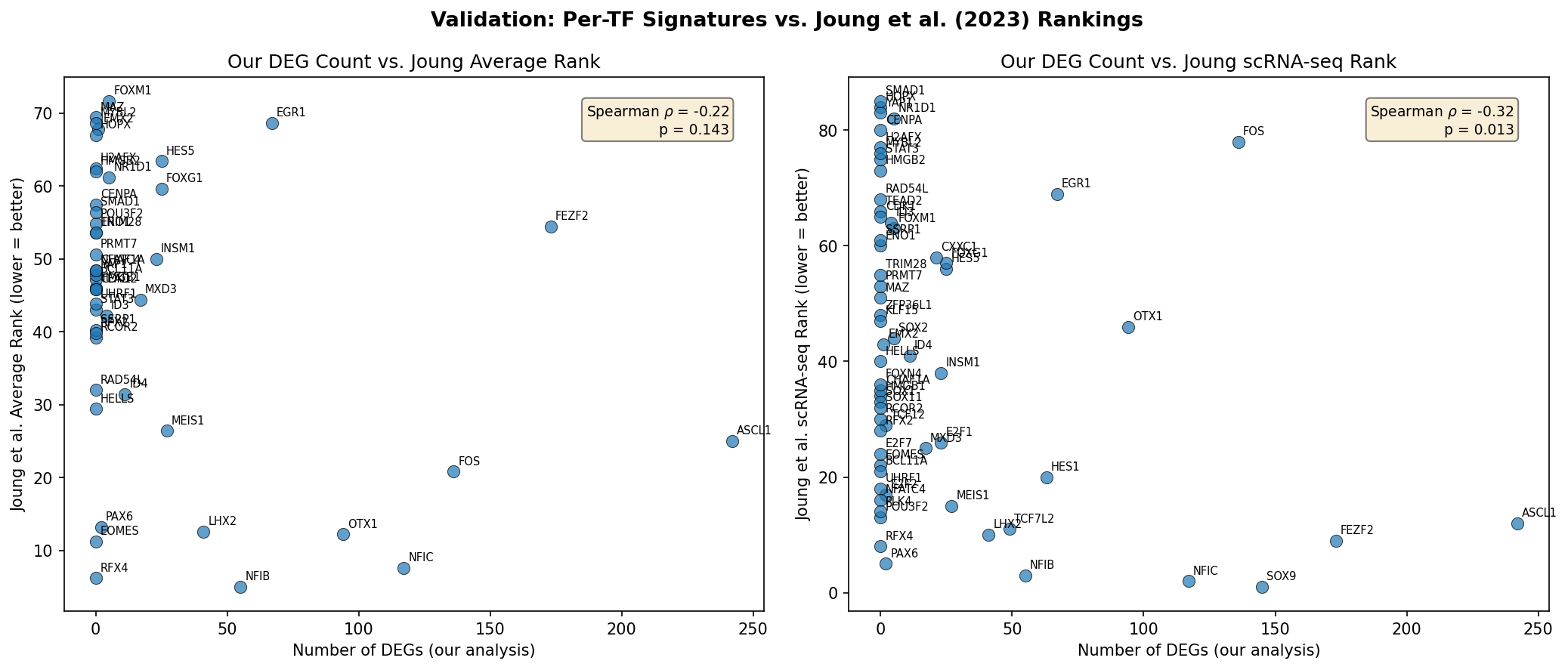}
  \caption{\textbf{Validation of per-TF signatures against Joung et al.\
    (2023) rankings.}
    Scatter plots comparing our per-TF DEG count with
    Joung et al.\ average rank (left) and scRNA-seq rank (right). Lower rank
    indicates stronger TF effect, so the negative Spearman $\rho$ reflects
    positive agreement between the two studies. The significant correlation
    ($\rho = -0.316$, $p = 0.013$ for scRNA-seq rank) validates that our
    independently derived signatures recapitulate the original study's findings.}
  \label{fig:joung_comparison}
\end{figure}

To validate our per-TF transcriptional signatures, we compared the number of
DEGs identified for each TF against the rankings published by Joung et al.\
in Table S1B \citep{joung2023transcription}. Of 61 TFs in our analysis, all
61 were present in the published rankings. We observed a significant negative
Spearman correlation between our DEG count and the Joung scRNA-seq rank
($\rho = -0.316$, $p = 0.013$; Figure~\ref{fig:joung_comparison}).
Because Joung et al.'s rankings are ordinal with rank~1 denoting the strongest
TF, a negative $\rho$ indicates \emph{positive agreement}: TFs with more DEGs
in our analysis receive lower (i.e., better) ranks in the original study.
The correlation with the Joung average rank
(combining arrayed, reporter, Flow-FISH, and scRNA-seq assays) showed a
consistent trend ($\rho = -0.219$, $p = 0.14$, $n = 46$ TFs with available
average ranks).

\section{Discussion}

Our re-analysis of the TF Atlas dataset demonstrates that the deposited
single-cell data, combined with the MORF barcode mapping, enables systematic
characterization of TF-specific transcriptional programs beyond the original
ranking analysis. Several key findings emerge:

\textbf{TF-specific signatures in pooled screens.} Using EB cells as a
negative control with background subtraction to remove shared
batch/transduction artifacts, 59 of 61 sufficiently represented TFs showed
TF-specific transcriptional signatures---a dramatic improvement over the 27
TFs detected with one-vs-rest alone. The background subtraction step (removing
1,946 genes DE in $\geq$43 of 61 TF groups) is critical: without it, raw
vs-EB comparisons are dominated by batch and transduction artifacts rather
than TF-specific signal. The strongest effects (HOPX, MAZ, PAX6, FOS, FEZF2)
correspond to TFs with established roles in cell fate determination. Our
signatures are validated by a significant correlation with the original Joung
et al.\ scRNA-seq rankings ($\rho = -0.316$, $p = 0.013$; negative $\rho$
reflects positive agreement, since lower rank = stronger TF), confirming that
our independent re-analysis recapitulates the published results.

\textbf{Missing negative controls.} The original MORF library includes GFP
and mCherry as negative controls (TFORF3549 and TFORF3550), and Joung et al.\
use them in their published analysis (Figure~2A). However, these controls are
absent from the deposited TFmap file, so we cannot identify control cells
within the pooled screen. We investigated whether the 7,451 undetected (NA)
cells could serve as untransfected controls, but their QC profile (median
2,768 genes, 10,443 UMIs) matches transduced cells rather than EB controls
(median 1,020 genes, 2,068 UMIs), indicating they are TF-expressing cells
with failed barcode detection. This necessitated our EB-based approach, which
effectively compensates for the missing intra-pool controls but requires the
background subtraction step to account for cross-batch confounds.

\textbf{Power considerations.} The comparison between condition-level
(thousands of DEGs) and per-TF pooled screen analysis highlights the
sensitivity trade-off in pooled designs. While pooled screens enable massively
parallel profiling, the per-TF cell counts (median 641 cells) limit
statistical power compared to dedicated overexpression experiments. The
EB-based approach with background subtraction substantially mitigates this:
by providing a proper untransfected baseline, it recovers TF-specific
signatures for 32 additional TFs that were undetectable with the one-vs-rest
design. Having the original GFP/mCherry controls within the pool would
further improve sensitivity by eliminating the need for cross-batch
comparisons.

\textbf{Pathway convergence and specificity.} At the condition level, multiple
TFs converge on shared pathways (Wnt, neurogenesis, Hippo), suggesting common
downstream effector programs in the embryoid body differentiation context. At
the per-TF level, enrichment analysis reveals both convergent signals (zinc/copper
ion response across 3--4 TFs) and highly specific signatures: FEZF2 uniquely
enriches for regulation of cell differentiation, NFIC for collagen biosynthesis,
and OTX1 for brain development, each consistent with established TF function.

\textbf{Limitations.} Our analysis uses the pre-computed TFmap for barcode
assignments rather than de novo demultiplexing from raw sequencing reads, as
the barcode library sequencing data was not deposited in GEO. The GFP and
mCherry negative controls from the original MORF library are absent from the
deposited TFmap, requiring the use of EB cells as an external control with
background subtraction to remove cross-batch artifacts. While this approach
successfully recovers TF-specific signatures for 59/61 TFs, having
intra-pool negative controls would provide a cleaner baseline without the
need for post-hoc artifact removal. Batch effects across the 8 pooled screen
replicates were assessed with Harmony and found to be minimal (3-iteration
convergence), so batch correction was not applied to the data used for DE.
Future work could apply more sophisticated methods such as
mixscape \citep{papalexi2021characterizing} or CINEMA-OT
\citep{dong2023causal} for improved perturbation effect estimation.

\textbf{Reproducibility.} Our complete pipeline---from raw data download
through enrichment analysis---is fully automated and idempotent, requiring
only a single command to reproduce all results from scratch. This
infrastructure enables the community to extend the analysis with additional
methods or gene set libraries.

\section{Conclusion}

We show that the deposited TF Atlas data can be converted into robust
TF-specific transcriptional signatures even though the original in-pool GFP and
mCherry controls are missing from the released barcode mapping. Using EB cells
as an external reference together with background subtraction, we recover
validated TF-specific signals for 59 of 61 testable TFs, far exceeding the 27
detectable by one-vs-rest alone. The resulting signatures reveal both shared
differentiation programs and TF-specific pathway effects, including FEZF2-linked
differentiation control and NFIC-linked collagen biosynthesis, while remaining
consistent with Joung et al.'s published TF rankings. More broadly, this work
provides a reproducible template for rescuing interpretable biology from large
public pooled perturbation datasets when deposited metadata is incomplete.

\section*{Supplementary Materials}

The following supplementary tables are available:
\begin{itemize}
  \item \textbf{Table S1}: Per-TF DEG summary from pooled screen analysis
    (61 TFs; cell counts, TF-specific DEG counts after background subtraction,
    one-vs-rest DEG counts, top up/downregulated genes).
  \item \textbf{Table S2}: Significant ORA enrichment terms per TF
    (207 terms across 13 TFs, adjusted $p < 0.05$).
  \item \textbf{Table S3}: Condition-level DEG summary (9 conditions vs.\ EB
    control; DEG counts and top genes).
  \item \textbf{Table S4}: Significant ORA enrichment terms per condition
    (3,801 terms across 9 conditions, adjusted $p < 0.05$).
\end{itemize}

All supplementary tables are provided as CSV files.

\section*{Acknowledgments}

This work used publicly available data from Joung et al.\ (2023), deposited
under GEO accession GSE216481. We thank the original authors for making their
data and barcode mapping publicly available.


\end{document}